%% file: 0_main.tex
\documentclass{article}
\usepackage{spconf}
\usepackage{subfigure}
\usepackage{amsmath,graphicx}
\usepackage{adjustbox}
\usepackage{amssymb,amsfonts}

\usepackage{algorithmic}
\usepackage{algorithm}
\usepackage{multirow}
\usepackage{url}

\usepackage{textcomp}

\DeclareMathOperator*{\argmin}{arg\,min}

\graphicspath{{./pics/}}

\title{Robust Dominant Periodicity Detection for Time Series\\ with Missing Data}
\name{Qingsong Wen\textsuperscript{\rm 1}, Linxiao Yang\textsuperscript{\rm 2}, Liang Sun\textsuperscript{\rm 1}}
\address{
\textsuperscript{\rm 1}DAMO Academy, Alibaba Group, 
Bellevue, USA\\
\textsuperscript{\rm 2}DAMO Academy, Alibaba Group, 
Hangzhou, China\\
}

\providecommand{\keywords}[1]
{
  \small	
  \textbf{\textit{Keywords---}} #1
}

\begin{document}
%\ninept
%
\maketitle
\begin{abstract}
Periodicity detection is an important task in time series analysis, but still a challenging problem due to the diverse characteristics of time series data like abrupt trend change, outlier, noise, and especially block missing data. In this paper, we propose a robust and effective periodicity detection algorithm for time series with block missing data. We first design a robust trend filter to remove the interference of complicated trend patterns under missing data. Then, we propose a robust autocorrelation function (ACF) that can handle missing values and outliers effectively. We rigorously prove that the proposed robust ACF can still work well when the length of the missing block is less than $1/3$ of the period length. Last, by combining the time-frequency information, our algorithm can generate the period length accurately. The experimental results demonstrate that our algorithm outperforms existing periodicity detection algorithms on real-world time series datasets.
\end{abstract}
%
%\begin{keywords}
\keywords{periodicity detection, seasonality detection, missing data, robust methods, time series}
%\end{keywords}
%

\input{1_introduction.tex}

\input{3_methodology.tex}

\input{4_experiment.tex}

\input{5_conclusion.tex}

\vfill\pagebreak

\clearpage
\bibliographystyle{IEEEbib.bst}
\bibliography{6_Dom_Period_bibfile}

\end{document}

%% file: 1_introduction.tex
\section{Introduction}

Many time series signals are characterized by repeating cycles, or periodicity in modern applications like the Internet of Things (IoT), Artificial Intelligence for IT Operations (AIOps), and cloud computing~\cite{kim2020periodicity,wen2022kddtimeseries,zhang2020anomaly}. Periodicity detection and adjustment are crucial in many real-world time series applications, like anomaly detection~\cite{jingkun20_TAD,tolas2021periodicity,zhang2022tfad}, forecasting~\cite{WenRobustPeriod20,xu2021two,zhou2022fedformer}, classification and clustering~\cite{hyndman2006:characterizeTS,vlachos2004identifying}, and decomposition~\cite{STL_cleveland1990stl,wen2019robuststl,yang2021robuststl}.
% time series similarity search~\cite{toyoda2013pattern}, 
% , and compression~\cite{multi-scale:compression:2009}. 
However, due to different and complicated characteristics of real-world time series data in IoT, AIOps, and cloud computing~\cite{wen2022kddtimeseries}, accurate periodicity detection remains a challenging problem. Here we briefly highlight three challenges for practical periodicity detection. Firstly, most time series are non-stationary, and the trend changes even abrupt trend changes commonly occur. 
% Secondly, the periodic patterns could be complicated and dynamic. 
Secondly, the time series data generally contains noises and outliers. 
Thirdly, many real-world time series data contain missing or even block missing values~\cite{yi2016st}. 

% faloutsos2019forecasting, WenRobustPeriod20

%Therefore, 
% Thus, how to conduct periodicity detection u deal with the time series data with a block of missing values remain crucial in practice. 
% Accurate periodicity detection is still challenging due to the complexity of real-world time series data. 
% forecasting~\cite{papadimitriou2003adaptive,faloutsos2019forecasting,iBTune19,theodosiou2011forecasting,prema2015time},

% Traditional two types of periodicity detecteion algorithms
% The periodicity detection problem has been researched for decades. 
The existing periodicity detection algorithms can be categorized into two groups: 1) frequency domain methods relying on periodogram after Fourier transform, such as Fisher's test~\cite{fisher1929tests,wichert2004identifying}; 2) time domain methods relying on autocorrelation function (ACF), such as methods in~\cite{Wang2006,Toller2019}.
To combine the advantages of both groups, recent joint time-frequency methods are proposed in~\cite{cPD_vlachos2005Autoperiod,almasri2011new,WenRobustPeriod20}. 
% Furthermore, the ensemble method which combines multiple periodicity detection methods together is also proposed to increase accuracy as in \cite{Toller2019}. 
However, the aforementioned methods cannot directly handle time series with missing data. 
In order to deal with missing values, the Lomb-Scargle periodogram based methods are proposed to detect periodicity as in \cite{hu2014periodicity,glynn2006detecting,lomb1976least}. 
% However, the Lomb-Scargle periodogram does not perform well in periodicity detection of time series with non-Gaussian noise and non-sinusoidal shapes~\cite{schimmel2001emphasizing}. 
Unfortunately, these methods cannot robustly address outliers and noises in time series.

% all the aforementioned challenges. 

% our contributions
% state our methodology, also discribe how to overcome the above mentioned challenges. 
In this paper, we propose a novel periodicity detection method to detect the dominant periodicity of time series under missing data robustly and accurately. Firstly, to mitigate the side effects introduced by trend, especially abrupt trend changes, we design a robust filter to remove the trend component under outliers and missing data. Next, to obtain accurate period length, we design a robust ACF module which can effectively deal with impulse random noise with unknown heavy-tailed noise, as well as block missing data. We rigorously prove that our algorithm can still work successfully when the length of the missing block is less than $1/3$ of the whole length of the time series. 
Lastly, we combine the time-frequency information for the final accurate period length estimation. Compared with various state-of-the-art algorithms, our proposed robust algorithm performs significantly better on real-world datasets, especially for time series with missing data.

%% file: 3_methodology.tex
\begin{figure*}[t]
\centering
    \includegraphics[width=0.7\linewidth]{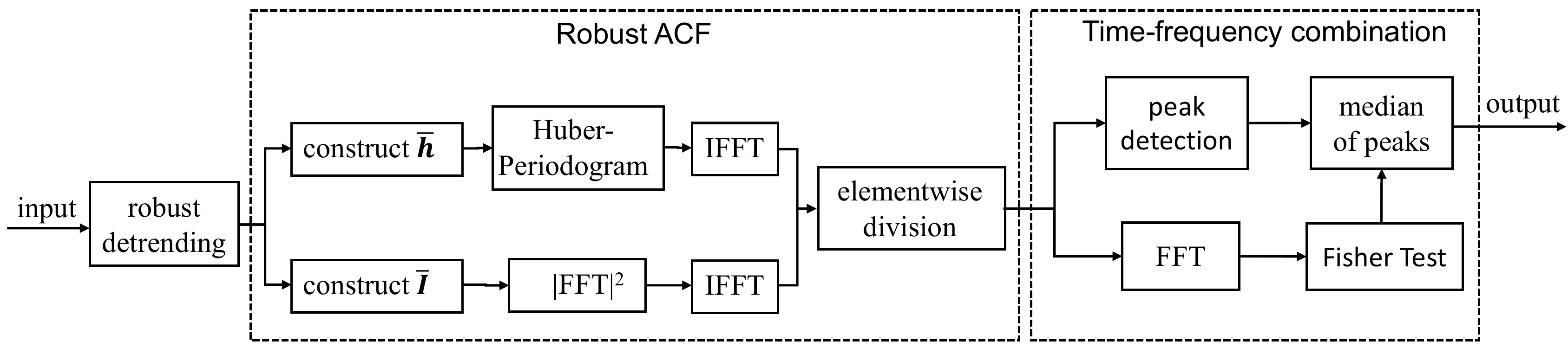}
    \vspace{-0.4cm}
    \caption{Framework of the proposed robust dominant periodicity detection algorithm for time series with missing data.}
    \vspace{-0.35cm}
\label{fig:Dominant_Periodicity_diagram}
\end{figure*}

\section{Proposed Periodicity Detection}

\subsection{Framework Overview}
We consider the following complex time series model with trend and dominant periodicity/seasonality as
\begin{equation}\label{eq:whole_model} 
y_t = \tau_t + s_t + r_t, \quad t = 0,1,\cdots,N-1
\end{equation}
where $y_t$ represents the observed value at time $t$, $\tau_t$ is the trend component, $s_t$ is the dominant periodic/seasonal component with period length $T$, and $r_t = a_t + n_t$ denotes the remainder part which contains the noise $n_t$ and possible outlier $a_t$. Note that $y_t$ may contain missing data. For dominant periodicity detection, we aim to identify if the time series is periodic and its major period length.

Our proposed periodicity detection algorithm contains three steps: 1) robust detrending filter; 2) robust ACF; 3) time-frequency combination.
The diagram of the proposed algorithm is depicted in Fig.~\ref{fig:Dominant_Periodicity_diagram}, which will be elaborated in the following subsections.

\vspace{-2mm}
\subsection{Robust Detrending under Missing Data}
\vspace{-1mm}
Most real-world time series usually have missing data with varying trend and outliers, 
% Here we first standardize the time series data to have a mean of 0 and a standard deviation of 1, which can remove the DC (direct current) component of the time series and facilitate the following FFT-based Periodogram analysis. 
which makes the ACF hard to capture the period information. 
To remove the trend component under missing data and motivated by the RobustTrend filter~\cite{wen2019robusttrend},
% to coarsely detrend time series, since it can capture both slow and abrupt trend changes while robust to outliers. Its trend extraction is to minimize the following objective function
% \begin{equation}\label{eq:robust_trend} 
% g_{\gamma} (\mathbf{y} - \boldsymbol{\tau}) 
% + \lambda_1 || \mathbf{D}^{(1)}\boldsymbol{\tau} ||_1
% + \lambda_2 || \mathbf{D}^{(2)}\boldsymbol{\tau} ||_1,
% \end{equation}
% where $g_{\gamma} (\mathbf{x})=\sum_i g_{\gamma} (x_i)$ is the summation of elementwise Huber loss function with each element as
% \begin{equation}\label{eq:huber_loss} 
%  g_{\gamma} (x_i) =
% \begin{cases} 
% \frac{1}{2} x_i^2,       & |x_i| \leq \gamma\\
% \gamma|x_i|- \frac{1}{2} \gamma^2, & |x_i| > \gamma
% \end{cases}
% \end{equation}
% And the ${D}^{(1)}\in \mathbb{R}^{(N-1)\times N}$ and ${D}^{(2)}\in \mathbb{R}^{(N-2)\times N}$ are the first-order and second-order difference matrix, i.e.,
% \begin{equation} \label{eq:Dk}
% \resizebox{1.0\hsize}{!}{$
% \mathbf{D}^{(1)} =  
% \begin{bmatrix}
% 1 & -1\\
%   & 1& -1\\
% & &\ddots\\
% &&1&-1
% \end{bmatrix},
% ~\mathbf{D}^{(2)} = 
% \begin{bmatrix}
% 1 & -2 &1&\\
% &1&-2&1&\\
% &&&\ddots\\
% &&&1&-2&1
% \end{bmatrix}.
% $}
% \end{equation}
% However, the original RobustTrend filter cannot work under missing data scenario. In order to overcome this limitation, 
we design a novel robust trend filter under missing data with objective function as
\begin{equation}\label{eq:modified_robust_trend} 
||\mathbf{W}(\mathbf{y} - \boldsymbol{\tau})||_1
+ \lambda_1 || \mathbf{D}^{(1)}\boldsymbol{\tau} ||_1
+ \lambda_2 || \mathbf{D}^{(2)}\boldsymbol{\tau} ||_1,
\end{equation}
where the ${D}^{(1)}\in \mathbb{R}^{(N-1)\times N}$ and ${D}^{(2)}\in \mathbb{R}^{(N-2)\times N}$ are the first- and second-order difference matrix adopted to capture abrupt and slow trend changes, respectively, with the following form:
\begin{equation} \label{eq:Dk}\notag
\resizebox{0.7\hsize}{!}{$
\mathbf{D}^{(1)} =  
\begin{bmatrix}
1 & -1\\
  & 1& -1\\
& &\ddots\\
&&1&-1
\end{bmatrix},
~\mathbf{D}^{(2)} = 
\begin{bmatrix}
1 & -2 &1&\\
&1&-2&1&\\
&&&\ddots\\
&&&1&-2&1
\end{bmatrix},
$}\vspace{-0.1cm}
\end{equation}
and $\mathbf{W} = \text{diag}\left(w_t\right)$ is a diagonal matrix with $w_t$ as 0 if $y_t$ is missing and 1 otherwise.
% \begin{equation}\label{eq:diag_W} 
% w_t =
% \begin{cases} 
% 0,  & \text{if $y_t$ is missing}\\
% 1, & \text{otherwise}
% \end{cases}
% \end{equation}
% By incorparting matrix $\mathbf{W}, we can perform trend extraction without missing value imputation. 
Note that our designed trend filter has two differences compared with the RobustTrend filter~\cite{wen2019robusttrend}. One is the $\mathbf{W}$ matrix incorporated into the loss function, which makes trend extraction possible without missing value imputation. The other is that
we update the Huber loss of the original RobustTrend filter to least absolute deviations (LAD) loss for its similar robustness to outliers and easy-to-solve property in alternative direction method of multipliers (ADMM). Specifically, to obtain the trend component from Eq.~\eqref{eq:modified_robust_trend}, we can rewrite it in an equivalent form:
\begin{equation}\label{eq:trend_LP_opt} 
\begin{matrix}
\min & 
||\mathbf{e}||_1 \\
\textrm{s.t.} & 
\mathbf{A}\boldsymbol{\tau}-\mathbf{b} = \mathbf{e}
\end{matrix}
\end{equation}
where $\mathbf{A} = [\mathbf{W}^T, \mathbf{D}^{(1)T}, \mathbf{D}^{(2)T}]^T$, $\mathbf{b}=[(\mathbf{Wy})^T, \mathbf{0}^T]^T$. 
% Then, the solution can be obtained by the ADMM procedure~\cite{boyd} 
% which is omitted here due to space limitation.
% \begin{equation}\label{eq:matrix_A_b}  %\notag
% \mathbf{A}= 
% \begin{bmatrix}
% \mathbf{\mathbf{W}}_{N \times N} \\
% \lambda_1 \mathbf{D}^{(1)}_{(N - 1) \times N}\\
% \lambda_2 \mathbf{D}^{(2)}_{(N - 2) \times N} \\
% \end{bmatrix}, 
% \mathbf{b}= 
% \begin{bmatrix}
% \mathbf{Wy}_{N \times 1}\\
% \mathbf{0}_{(2N - 3) \times 1}\\
% \end{bmatrix}.
% \end{equation} 
Then we can obtain the augmented Lagrangian as\vspace{-2mm}
\begin{equation}\label{eq:Lagrangian} \notag
L_{\rho}(\mathbf{x},\mathbf{y},\mathbf{u}) = 
 || \mathbf{y} ||_1 +
\mathbf{u}^T (\mathbf{Ax}-\mathbf{b}-\mathbf{y}) + \frac{\rho}{2} ||\mathbf{Ax}-\mathbf{b}-\mathbf{y}||_2^2\vspace{-2mm}
\end{equation}
where $\mathbf{u}$ is the dual variable, and $\rho$ is the penalty parameter.
Next, the solution can be obtained through the ADMM~\cite{boyd} as\vspace{-1mm}
% by updating the following steps as
\begin{align} \notag
\mathbf{x}_{t+1} \! &=\! \argmin_{\mathbf{x}} L_{\rho}(\mathbf{x},\mathbf{y},\mathbf{u}) \! =\! (\mathbf{A}^T \!\! \mathbf{A})^{-1}\! \mathbf{A}^T \!(\mathbf{b} + \mathbf{y}_t \!-\! \mathbf{u}_t/{\rho})\\ \notag%\label{eq:admm_xx}   \notag
\mathbf{y}_{t+1} \! &=\! \argmin_{\mathbf{y}} L_{\rho}(\mathbf{x},\mathbf{y},\mathbf{u}) \! =\! {S}_{1/\rho} (\mathbf{Ax}_{t+1} - \mathbf{b} + \mathbf{u}_t/\rho)\\  \notag % \label{eq:admm_zz} 
\mathbf{u}_{t+1} \! &=\! \argmin_{\mathbf{u}} L_{\rho}(\mathbf{x},\mathbf{y},\mathbf{u}) \! =\!  \mathbf{u}_{t} + {\rho}(\mathbf{Ax}_{t+1} - \mathbf{y}_{t+1} - \mathbf{b}) %\label{eq:admm_uu}
\end{align} 
where ${S}_{\kappa}(x)$ is the soft thresholding as ${S}_{\kappa}(x)=(1-\kappa/|x|)_{+}x$.

% Note that both periodogram and ACF are prone to outliers, and leading to inaccurate periodicity detection. To handle outliers properly, we adopt Hampel filter \cite{ACFPSD_pearson2016Hampel}: 
% \begin{equation}\label{eq:Hample_filter} 
% x_t =
% \begin{cases} 
% y_t',  & |y_t'-m| \leq \gamma S\\
% m, & |y_t'-m|> \gamma S
% \end{cases}
% \end{equation}
% where $x_t$ is the response of Hampel filter, $\gamma$ is a threshold parameter with a typical value as 3, $m$ denotes the median value, and $S$ is the Mean Absolute Deviation (MAD) which is defined as
% \begin{equation}\label{eq:median_MAD} 
% \quad S=1.4826 \times median|y_t'-m|.
% \end{equation}
% The factor 1.4826 in Eq.~\eqref{eq:median_MAD} makes the MAD an unbiased estimate of the standard deviation in case of Gaussian data.

% (TODO)
% RobustTrend filter,
% Method for trend filtering when there are missing data?
% 1, just linear interpolation.

% 2,
% paper--"Trend Extraction From Time Series With Structural Breaks and Missing Observations"

% "The major distinction
% to be drawn here is that the linearly interpolated points are used
% as supplemental training data for the regression algorithm, as
% opposed to being the predicted values in and of themselves."

% A common problem encountered in real time series analysis is missing data.

\input{3_z_1_MissingDataProcess}

% Then we can obtain the augmented Lagrangian as
% \begin{equation}\label{eq:Lagrangian} \notag
% L_{\rho}(\boldsymbol{\beta},\mathbf{z},\mathbf{v}) = 
% \gamma_{\zeta}^{hub}(\mathbf{z}) +
% \mathbf{v}^T ( \boldsymbol{\phi} \boldsymbol{\beta} -  \mathbf{x}  - \mathbf{z}) + \frac{\rho}{2} || \boldsymbol{\phi} \boldsymbol{\beta} -  \mathbf{x}  - \mathbf{z}||_2^2
% \end{equation}
% where $\mathbf{v}$ is the dual variable, and $\rho$ is the penalty parameter.
% Following the ADMM procedure~\cite{boyd2011distributed} and taking proximal operator for huber function $\gamma_{\zeta}^{hub} ( \mathbf{z})$~\cite{Parikh2014}, we can obtain the updating steps as
% \begin{align} % \label{eq:admm}
% \boldsymbol{\beta}^{k+1} ={}&  (\boldsymbol{\phi}^T \boldsymbol{\phi} )^{-1} \boldsymbol{\phi}^T (\mathbf{z}^k+\mathbf{x}-\mathbf{u}^k)\label{eq:admm1}\\
% \begin{split} 
% \mathbf{z}^{k+1} ={}& \frac{\rho}{1+\rho}\left( \boldsymbol{\phi} \boldsymbol{\beta}^{k+1} + \mathbf{u}^{k}  - \mathbf{x} \right) +  \\
%  & \frac{1}{1+\rho}S_{\frac{\zeta(1+\rho)}{\rho}}\left( \boldsymbol{\phi} \boldsymbol{\beta}^{k+1} + \mathbf{u}^{k}  - \mathbf{x}  \right) 
% \end{split}\label{eq:admm2}\\
% \mathbf{u}^{k+1} ={}& \mathbf{u}^{k} +  \boldsymbol{\phi} \boldsymbol{\beta}^{k+1} - \mathbf{z}^{k+1}- \mathbf{x}  \label{eq:admm3}
% \end{align}
% where $\mathbf{u}=(1/\rho)\mathbf{v}$ is the scaled dual variable to make the formulation more convenient.
% The soft thresholding operation ${S}_{\rho}(x)$ in $\mathbf{z}-$minimization step can be efficiently calculated by ${S}_{\rho}(x) = (1-\rho/|x|)_{+}x$.

\vspace{-2mm}
\subsection{Final Time-Frequency Combination}\label{sec:RobustFisher}
\vspace{-2mm}
% We consider both time-domain and frequency-domain information to increase the accuracy of periodicity detection. 

In frequency domain, to detect dominant periodicity, Fisher's test \cite{fisher1929tests} defines $g$-statistic as
$
g = {\text{max}_{k}{P_k}}/{\sum_{j=1}^{N} \!\!P_j}, k=1,2, \cdots, N,
$
% \begin{equation}\label{eq:Fisher_g-statistics} 
% g = \frac{\text{max}_{k}{P_k}}{\sum_{j=1}^{N}P_j}, \quad j=1,2, \cdots, N,
% \end{equation}
where $P_k=\text{FFT}\{r_t\}$ is the calculated periodogram via Wiener-Khinchin theorem~\cite{Wiener1930} based on robust ACF $r_t$. Therefore, this $g$-statistic is also robust to outliers and missing data. 
The distribution of $g$-statistic gives a $p$-value to determine if a time series is periodic~\cite{fisher1929tests}.
If this value is less than the predefined threshold $\alpha$, we reject the null hypothesis $H_0$ and conclude the time series is periodic with period length candidate as $N/k$ where $k = \arg \max_k P_k$.
If the Fisher's test is passed, 
we further refine the candidate of period length in the time domain similar to~\cite {cPD_vlachos2005Autoperiod,WenRobustPeriod20}. Specifically, we first summarize the peaks of robust ACF through 
peak detection~\cite{Scholkmann2012}. Then, we calculate the median distance of those peaks whose heights exceed predefined threshold. Furthermore, based on the resolution of periodogram, i.e., the peak value of $P_k$ at index $k$ corresponds to period length in the range $[\frac{N}{k}, \frac{N}{k-1})$, the median distance of ACF peaks is the final period length only if it locates in 
$R_k = \left[\frac{1}{2}\left(\frac{N}{k+1} + \frac{N}{k}\right)-1, \cdots, \frac{1}{2}\left(\frac{N}{k} + \frac{N}{k-1}\right)+1\right]$.
% \begin{equation}\notag%\label{eq:ACFrange} 
% R_k = \left[\frac{1}{2}\left(\frac{N}{k+1} + \frac{N}{k}\right)-1, \cdots, \frac{1}{2}\left(\frac{N}{k} + \frac{N}{k-1}\right)+1\right].
% \end{equation}
Note that this combination is necessary. On the one hand, the periodogram has limited resolution and spectral leakage may exist~\cite{cPD_vlachos2005Autoperiod}, which makes the candidate from Fisher's test not accurate. On the other hand, only relying on ACF may result in false positive results since ACF cannot provide if there is dominant periodicity.

%% file: 3_z_1_MissingDataProcess.tex
\vspace{-2mm}
\subsection{Roubust ACF with M-Periodogram}
\vspace{-1mm}
\subsubsection{Structure of the Proposed Roubust ACF}
% \vspace{-0.1cm}
Let $\{x_t\}$ denote the detrended time series, i.e. $x_t=y_t-\tau_t$. Then
its period can be estimated by finding the position of the largest value
of the ACF $\{r_k\}$
which is defined as $r_k=\mathbb{E}(x_{t+k}x_t)$. 
The ACF
is usually estimated using
$r_k\approx\frac{1}{|\mathbb{Q}_k|}\sum_{t\in \mathbb{Q}_k}x_{t+k}x_t,
$
where set $\mathbb{Q}_k$ is defined as $\{t|0\le t\le N-1, 0\le n+k\le N-1\}$,
and $|\mathbb{Q}_k|$ denote the size of $\mathbb{Q}_k$.
However, this estimator can not be applied to time series with missing values directly. To address this problem, we propose a new unbias estimator.
To better illustrate our method, let $\{I_t\}$ be a binary
sequence that indicates whether $\{x_t\}$ is observed.
Specifically, let $I_t=1$ when $x_t$ is observed, and
$I_t=0$ when $x_t$ is missing.
Our proposed ACF estimator is
$r_k\approx\frac{1}{|\hat{\mathbb{Q}}_k|}\sum_{t\in \hat{\mathbb{Q}}_k} x_{t+k}x_t
$
where the set $\hat{\mathbb{Q}}_k=\{t|I_t=1, I_{t+k}=1\}$.
It is ready to see that this estimator is an 
unbiased estimator of the autocorrelation, but directly computing it has a computational
complexity of order $O(N^2)$. To reduce the complexity, we define a sequence $\{h_t\}$
by padding $\{x_t\}$ with zeros, i.e. $h_t=x_t$ if $I_t=1$ and $h_t=0$ if $I_t=0$.
% \begin{align}
%     h_t=
%     \begin{cases}
%     x_t&I_t=1\\
%     0&I_t=0
%     \end{cases}
% \end{align}
It is easy to see that $h_t=x_t I_t$.
Then our proposed ACF estimator can be rewritten as
\begin{align}\label{estimator2}
    r_k\approx&\frac{1}{|\hat{\mathbb{Q}}_k|}\sum_{t\in \hat{\mathbb{Q}}_k} x_{t+k}x_t
    =\frac{1}{|\hat{\mathbb{Q}}_k|}\sum_{t\in \hat{\mathbb{Q}}_k} I_{t+k}x_{t+k}I_t x_t\nonumber\\
    % =&\frac{1}{|\hat{\mathbb{Q}}_k|}\sum_{t\in \hat{\mathbb{Q}}_k} h_{t+k}h_t
    % =\frac{1}{|\hat{\mathbb{Q}}_k|}\sum h_{t+k}h_t\nonumber\\
    =&\frac{1}{|\hat{\mathbb{Q}}_k|}\sum h_{t+k}\hat{h}_{k-(t+k)}
    =\frac{1}{|\hat{\mathbb{Q}}_k|}(\mathbf{h}*\hat{\mathbf{h}})_k
\end{align}
where $\hat{h}_t=h_{-t}$, and the operation ``*'' denotes the linear 
convolution. Recall that $\hat{\mathbb{Q}}_k=\{t|I_t=1, I_{t+k}=1\}=\{t|I_t I_{t+k}=1\}
$, we have 
$|\hat{\mathbb{Q}}_k|=\sum I_n I_{t+k}
=\sum \hat{I}_{-t}I_{t+k}
=(\hat{\mathbf{I}}*\mathbf{I})_k$, where $\hat{I}_t=I_{-t}$. Furthermore, we can utilize FFT/IFFT based on circular convolution theorem to reduce the complexity of $(\mathbf{h}*\hat{\mathbf{h}})_k$ and $(\hat{\mathbf{I}}*\mathbf{I})_k$ from $O(N^2)$ to $O(N\log N)$. Therefore, our efficient and robust ACF is 
\begin{align}\label{eq:estimator3}
    r_k\approx \frac{(\mathbf{h}*\hat{\mathbf{h}})_k}{(\hat{\mathbf{I}}*\mathbf{I})_k}
    = \frac{\text{IFFT}(|\text{FFT}(\bar{\mathbf{h}})|^2)_k}{\text{IFFT}(|\text{FFT}(\bar{\mathbf{I}})|^2)_k},
\end{align}
where the length of $\mathbf{h}$ and $\mathbf{I}$ are doubled by padding $N$ zeros and denoted as $\bar{\mathbf{h}}=[{\mathbf{h}}^T, 0, \cdots, 0]^T$ and $\bar{\mathbf{I}}=[{\mathbf{I}}^T, 0, \cdots, 0]^T$.

Besides dealing with missing data, we also consider to mitigate the effect of outliers in $\bar{\mathbf{h}}$ of Eq.~\eqref{eq:estimator3}. Note that $|\text{FFT}(\bar{\mathbf{h}})|^2$ is the conventional periodogram of $\bar{\mathbf{h}}$.
% with each element as
% \begin{equation}\label{eq:Periodogram11} 
% (|\text{FFT}(\bar{\mathbf{h}})|^2)_k = \frac{1}{{N'}} \sum_{t=0}^{N'-1}  \left|\left| \bar{h}_te^{-i2 \pi kt/N'}\right|\right|^2,
% \end{equation}
% where $N'=2N$.
To make it robust to outliers, we introduce 
% the M-periodogram~\cite{katkovnik1998robust}. Specifically, we adopt 
the Huber-periodogram~\cite{WenRobustPeriod20} as %~\cite{wen2020robustperiod}
\begin{equation}\label{eq:Periodogram_final} %\notag
(|\text{FFT}(\bar{\mathbf{h}})|^2)_k \approx \frac{N'}{4} \left|\boldsymbol{\hat{\beta}}_{M}(k)\right|^2
= \frac{N'}{4} \left|   \underset{\boldsymbol{\beta} \in \boldsymbol{R}^2}{\arg\min}
~\!\gamma( \boldsymbol{\phi} \boldsymbol{\beta} -  \bar{\mathbf{h}} ) \right|^2,
\end{equation}
where $N'=2N$, $\boldsymbol{\hat{\beta}}_{M}(k)$ is the robust estimation of harmonic regressor $\boldsymbol{\phi}_t  = [\cos(2 \pi kt/N'),\sin(2 \pi kt/N')]$,
$\boldsymbol{\phi}$ has the matrix forms of
$[\boldsymbol{\phi}_0,\boldsymbol{\phi}_1, \cdots, \boldsymbol{\phi}_{N'-1}]^T$, and $\gamma(\mathbf{x})$ is the robustifying objective function defined as the summation of Huber loss as $\gamma( \mathbf{x}) = \sum_{t=0}^{N'-1} \gamma^{hub}(x_t) $.

Based on the aforementioned processing for missing data and outliers, the final structure of our robust ACF is depicted in the middle part of Fig.~\ref{fig:Dominant_Periodicity_diagram}.

\subsubsection{Theoretical Constraint under Missing Data}
\vspace{-2mm}
Next, we investigate the maximum missing block length that our robust ACF estimator can bear. 
Clearly, utilizing Eq.~(\ref{estimator2}) to compute the ACF
requires $Q_k=|\hat{\mathbb{Q}}_k|>0$. While the value of $Q_k$
depends on the positions and the volume of the missing entries,
which makes it difficult for analysis. In practice, the worst scenarios for missing
entries occur in cluster. In the following, we provide
theoretical analysis for the scenario that all the missing entries concentrate in one block, i.e., $I_k=0$ when $m\le n \le m+l-1$ and $I_k=1$ otherwise, 
% \begin{align}
%     I_k=
%     \begin{cases}
%     0 & m\le n \le m+l-1\\
%     1 & \text{else}
%     \end{cases}
% \end{align}
where $m$ denotes the start position of the block and
$l$ denotes the length of the block.
Here we assume that the block does not include the beginning and
end point of the sequence, i.e.,
$m>0$ and $m+l-1<N-1$.
Note that the proposed method can work if and only if $Q_k>0$. Recall that $Q_k$ is $\sum_{t=k}^{N-1} I_t I_{t-k}$ when $k\ge0$ and $\sum_{t=0}^{N+k} I_t I_{t-k}$ otherwise.
% \begin{align}
%     Q_k=\begin{cases}
%     \sum_{t=k}^{N-1} I_t I_{t-k}& k\ge0\\
%     \sum_{t=0}^{N+k} I_t I_{t-k}& k<0
%     \end{cases}
% \end{align}
Formally, we have the following proposition.

\textit{Proposition:} For any $k$ and $m$ that satisfy
$m>0$ and $m+l-1<N-1$, if $l<\frac{N}{3}$, we always have
$Q_k>0$.

% assume
% all the missing entries occurs consequently, i.e., the missing
% entries form a block. we investigate how long the block is that 
% our method can bear. Suppose we have a sequence of length $N$, whose
% indicator sequence is denoted by $\{I_n\}_{n=1}^{N}$.
% As shown above, the proposed method can work only if $Q_k>0$
% for all $-N+1\le k\le N-1$, where
% \begin{align}
%     Q_k=\begin{cases}
%     \sum_{n=k+1}^N I[n]I[n-k]& k\ge0\\
%     \sum_{n=1}^{N+k} I[n]I[n-k]& k<0
%     \end{cases}
% \end{align}

% Suppose all the missing entries occurs in one block that of length $l$,
% mathematically, we assume
% \begin{align}
%     I[n]=
%     \begin{cases}
%     0 & m\le n \le m+l-1\\
%     1 & \text{else}
%     \end{cases}
% \end{align}
% The we have following proposition

% \textit{Proposition 1:} For any $1<m<N-l+1$, if $l<\frac{N}{3}$, we always have
% $Q_k>0$.

\textit{Proof:} From the definition of $Q_k$, we have $Q_k\ge0$, then we only need to prove that it does not exist a $k$ such that $Q_k= 0$. Without loss of generality, we only consider
the case that $k>0$. Let $g_t=I_{t-k}$, then $g_t=0$ for $m+k\le t\le m+l+k-1$.
Suppose there is a $k$ and $m$ such that $Q_k=0$, then
$I_n\hat{I}_n=0$ for all $n\in [k,N-1]$, then we have\vspace{-1mm}
\begin{align}
    [k,N-1] \subset [m,m+l-1] \cup [m+k,m+l+k-1] \label{condition}
\end{align}\vspace{-1mm}
Note that Eq.~(\ref{condition}) implies the following three possibilities:
\begin{equation}\vspace{-2mm}
    [k,N-1] \subset [m,m+l-1]\label{case1}
\end{equation}
% \vspace{-3mm}
\begin{equation}\vspace{-2mm}
    [k,N-1] \subset [m+k,m+l+k-1]\label{case2}
\end{equation}
\begin{align}\vspace{-2mm}
    \!& [k,N-1] \subset [m,m+l-1]\cup [m+k,m\!+\!l\!+\!k\!-\!1]\nonumber\\
    \!& [m,\!m\!+\!l\!-\!1]\!\cup\! [m\!+\!k,\!m\!+\!l\!+\!k\!-\!1] \!=\! [m,\!m\!+\!l\!+\!k\!-\!1]\label{case3}
\end{align}
Obviously, Eqs.~(\ref{case1}) and (\ref{case2}) requires $N-1\le m+l-1$ or $k+1\ge m+k$
which is contradicted to $m<N-l$ and $m>0$.
Note that Eq.~(\ref{case3}) implies
$m\le k$, $m+k\le m+l$, and $N-1\le m+l+k-1$,
% \begin{align}
%     & m\le k\\
%     & m+k\le m+l\\
%     & N-1\le m+l+k-1
% \end{align}
which can be rewritten as\vspace{-1mm}
\begin{align} 
   \max(m,N-m-l)\le k \le l\label{condition2} 
\end{align}\vspace{-1mm}
We show that if $l<\frac{N}{3}$ , one cannot find a $m$
and $k$ satisfying Eq.~(\ref{condition2}).
It is easy to see that 
for all $0< m < N-l$, the minimum of $\max(m,N-m-l)$
is archived at $m=\frac{N-l}{2}$ and equals to 
$\frac{N-l}{2}$. As $l<\frac{N}{3}$,
we can see that $l$ is always less than $\frac{N-l}{2}$,
i.e.,
$l<\frac{N-l}{2}\le\max(m,N-m-l)
$
which makes there is no $k$ satisfying Eq.~(\ref{condition2}).
The proof is completed here.

%% file: 4_experiment.tex
\begin{figure}[!t]
    \centering
    % \vspace{-0.4cm}
    \subfigure[One CRAN periodic time series and its variant with outliers and missing values.]{
        \includegraphics[width=0.475\linewidth]{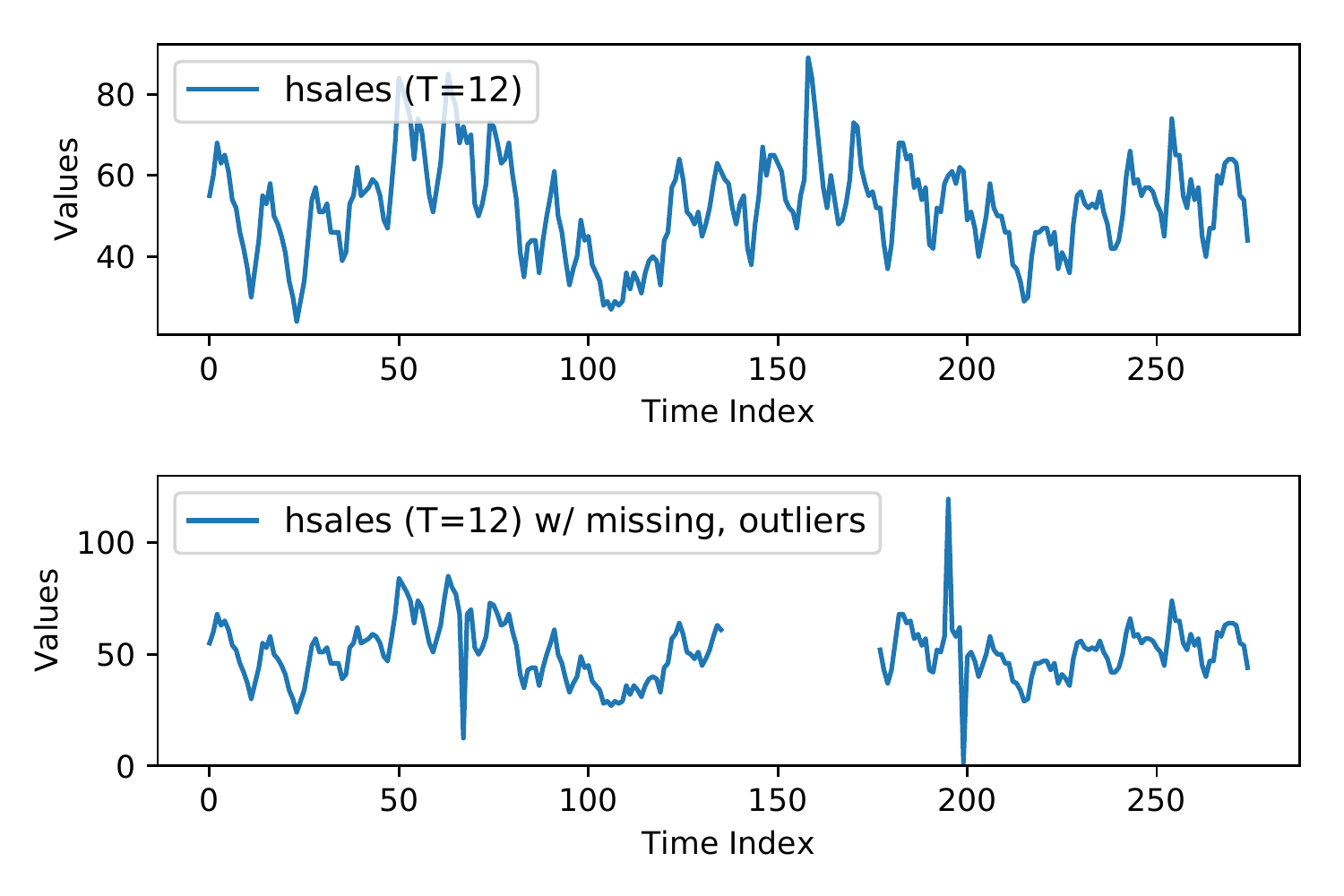}
        \label{fig:CRAN_data}
        }
        % \\ \vspace{-0.2cm}\
    \subfigure[Three typical periodic time series from cloud computing industry with outliers and missing values.]{
        \includegraphics[width=0.47\linewidth]{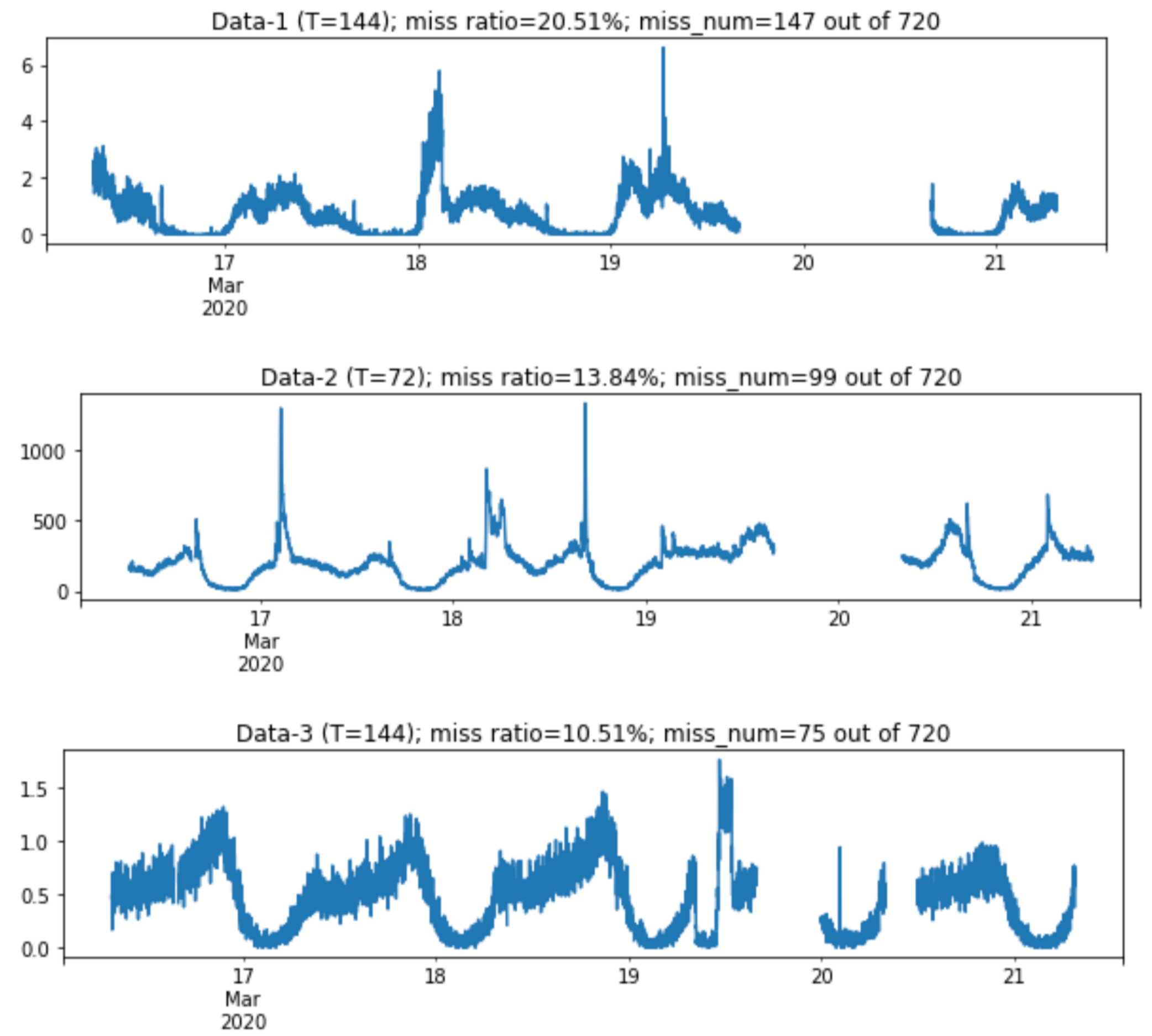}
        \label{fig:real_data}
        } 
        \vspace{-0.5cm}
    \caption{{Representative challenging periodic time series.}}
    \vspace{-0.5cm}
\end{figure}
% \clearpage

\vspace{-1mm}
\section{Experiments and Discussions}
\vspace{-1mm}
% In this section, we study the proposed robust periodicity detection algorithm empirically in comparison with other recent periodicity detection algorithms on both challenging synthetic and real-world datasets under missing data. We also investigate how each component in our algorithm works robustly under missing data.

% We conduct experiments on synthetic datasets and public datasets to demonstrate the effectiveness of the proposed RobustPeriod algorithm for periodicity detection and estimation.

% \vspace{-.3cm}

\vspace{-1mm}
\subsection{Periodicity Detection Comparisons}
% \subsubsection{Baseline Algorithms and Datasets}
\vspace{-2mm}
{\hspace{-0mm}\bf Baseline Algorithms and Datasets:}\\
We compare our algorithm with baseline algorithms: ACF-Med (median distance of ACF peaks), Fisher's Test, Lomb–Scargle periodogram~\cite{hu2014periodicity,glynn2006detecting,lomb1976least}, and the state-of-the-art time-frequency method RobustPeriod~\cite{WenRobustPeriod20}. Note that there exist other types of time-frequency algorithms~\cite{cPD_vlachos2005Autoperiod,almasri2011new}, but their performances are inferior to RobustPeriod~\cite{WenRobustPeriod20}. 
Except Lomb–Scargle and our proposed algorithm, we add linear interpolation in case of missing data for other algorithms since their performances are degraded severely if directly working on time series with missing values.

For datasets, we consider the public single-period time series from CRAN~\cite{cran_hornik2012comprehensive,Toller2019}, which contains 82 real-world time series in various scenarios (like IoT, sales, sunspots, etc.) with lengths from 16 to 3024 and period lengths from 2 to 52. We also add outliers and block missing data in the datasets to evaluate the robustness of all algorithms. 
The outlier ratio (OR) is set to 0.01 or 0.05 with outlier amplitude as 5 times standard deviation of original time series, and the block missing ratio (MR) is set to 0.05 or 0.30.
Since the outlier ratio is usually small in practice, we set it as 0.05 at most; while the missing ratio could be relatively large due to block missing, we set it as 0.3 at most.
One example is shown in Fig.~\ref{fig:CRAN_data}. Besides, We utilize 3 typical real-world datasets from a leading cloud computing company as shown in Fig.~\ref{fig:real_data}. The length of the dataset is 5 days with sampling resolution of 10 minutes, where their true period lengths are 144 (one-day), 72 (half-day), and 144 (one-day), respectively. For the evaluation, the precision is calculated by the ratio of the number of time series with correctly estimated period length to the total number of time series.

\begin{table*}[!t] % \begin{table*}[t]
\centering
% \footnotesize%\small
\caption{{Precision comparisons of different periodicity detection algorithms on public CRAN data. MR and OR indicate missing ratio and outlier ratio, respectively. The best results are highlighted in bold.}}
\label{tab:single_period_det}
% \vspace{-4mm}
\scalebox{0.7}{
\begin{tabular}{c|c|c|c|c|c|c|c|c|c}
\hline
\multirow{3}{*}{Algorithms} 
 & \multicolumn{9}{c}{CRAN dataset with or without missing values \& outliers}    \\ \cline{2-10}
                 &  \multicolumn{3}{c|}{MR=0}      & \multicolumn{3}{c|}{MR=0.05}  & \multicolumn{3}{c}{MR=0.30}  \\  \cline{2-10} 
                 &   \multicolumn{1}{c|}{OR=0}    & \multicolumn{1}{c|}{OR=0.01} & \multicolumn{1}{c|}{OR=0.05} &   \multicolumn{1}{c|}{OR=0}    & \multicolumn{1}{c|}{OR=0.01} & \multicolumn{1}{c|}{OR=0.05} &   \multicolumn{1}{c|}{OR=0}    & \multicolumn{1}{c|}{OR=0.01} & \multicolumn{1}{c}{OR=0.05}\\  
\hline\hline
ACF-Med        &  0.57  &      0.55     &  0.24  &  0.56  &  0.50  &  0.23  &  0.52 &  0.50  &  0.27 \\ \hline
Fisher's Test  &  0.52  &      0.44     &  0.32  &  0.49  &  0.44  &  0.30  &  0.42 &  0.33   &  0.24   \\ \hline
% AUTOPEIROD     &  0.57  &      0.46     &  0.21  &  0.57  &  0.45  &  0.23  &  0.61  &  0.65  &  0.30 \\ \hline
Lomb–Scargle   & 0.45   &      0.13     &  0.09  &  0.45  &  0.13  &  0.10  &  0.37  &  0.11   &  0.10 \\ \hline
RobustPeriod   &  0.62  &      0.60     &  0.59  &  0.59  &  0.57  &  0.50  &  0.47 &  0.45   &  0.43 \\ \hline
\textbf{Proposed}&\textbf{0.63}&    \textbf{0.62}     &  \textbf{0.60}  &  \textbf{0.60}  &  \textbf{0.60} &  \textbf{0.59}  &  \textbf{0.56}  &  \textbf{0.52}  &  \textbf{0.51} \\ \hline
% SAZED$_{opt}$  & 0.55        &       0.42    & 0.15   \\ \hline TODO in future
\end{tabular}
}
\vspace{-5mm}
\end{table*}

\begin{table}[t]%[!htbp]%[!htb]
% \small% \footnotesize%\small
\centering
% \tiny, \scriptsize, \footnotesize, \small, \normalsize, \large, \Large, \LARGE, \huge, and \Huge.
\caption{{Comparisons of different periodicity detection algorithms on 3 typical real-world datasets with missing data.}}
% \vspace{-4mm}
\label{tab:realData_diff_algs}
% \begin{adjustbox}{max width=0.47\textwidth}
\scalebox{0.8}{
\begin{tabular}{c|c|c|c}
\hline
% Algorithms & Data-2, \textbf{T=144} & Data-3, \textbf{T=288} & Data-4, \textbf{T=(6,12)} & Data-5, \textbf{T=(48,336)} \\\hline \hline
Algorithms & Data-1 (T=144) & Data-2 (T=72) & Data-3 (T=144) \\\hline \hline
% Huber-ACF-Med  & 144 & 288 & 6 & 48 \\ \hline
ACF-Med  & 144 & 72 & 143 \\ \hline
Fisher's Test  & 144  & 73  &  143  \\ \hline
% AUTOPERIOD & (148,288)  & (73,154) & 143  \\ \hline
Lomb–Scargle  & 140 & 71 & 140  \\ \hline
RobustPeriod &  142 &  (72,144) & 144  \\ \hline
\textbf{Proposed}  & \textbf{144} & \textbf{72} & \textbf{144}  \\ \hline
% Ground truth & {144} & 72 & 144 \\ \hline
\end{tabular}
}
% \end{adjustbox}
\vspace{-5mm}
\end{table}

% \vspace{-.2cm}
% \subsubsection{Performance Comparisons of Different Algorithms}
\vspace{1.5mm}
{\hspace{-3.5mm}\bf Performance Comparisons of Different Algorithms:}\\
% \vspace{-.2cm}
We summarize the detection precision results on public CRAN datasets with or without missing data and outliers in Table~\ref{tab:single_period_det}. It can be seen that, when there are no missing data and outliers, both  RobustPeriod and our proposed algorithm exhibit similar results and have better performance than others. 
When there are no outliers but with increasing missing data (i.e., OR=0, MR=0, 0.05, 0.3), the performance degradation of Lomb–Scargle is relatively smaller than ACF-Med, Fisher's Test, and RobustPeriod. This is due that Lomb–Scargle can directly deal with missing data without imputation. However, the Lomb–Scargle is not robust to outliers, and its performance becomes much worse than others when there are outliers. 
When we fix the ratio of missing data but with increasing outliers (i.e., OR=0, 0.05, 0.3), the performance degradation of RobustPeriod is much smaller than ACF-Med, Fisher's Test, and Lomb–Scargle, which demonstrates its robustness to outliers. 
When there are both missing data and outliers, 
the performances of existing algorithms drop significantly, especially for ACF-Med, Fisher's Test, and Lomb–Scargle methods. The reason is that these methods heavily depend on conventional ACF or FFT while the results of ACF or FFT would be severely distorted by block missing data and outliers.
Overall, our proposed algorithm performs best in all test cases, especially exhibits better performance than others when the missing data and outliers ratio are severe.

Table~\ref{tab:realData_diff_algs} summarizes the detection results of the 3 typical real-world datasets as in Fig.~\ref{fig:real_data}.  It can be seen that the RobustPeriod may generate some false positives, which is due to the fact that it is designed for general multiple-period detection. It is also interesting to find that Lomb–Scargle is not robust to large outliers and noises even though it can directly deal with missing data. Overall, our proposed algorithm achieves the best results.

\vspace{-3mm}
\subsection{Component Investigation}
\vspace{-2mm}

% \subsubsection{Trend Extraction with Outliers and Missing Data}
% \begin{figure}[!t]
%     \centering
%     \subfigure[Trend extraction without missing data and outliers. ]{\includegraphics[width=0.35\textwidth]{plots/detrend_ori.png}\label{fig:detrend_wo_missing}}\\
%     \subfigure[Trend extraction with missing data and outliers.]{\includegraphics[width=0.35\textwidth]{plots/detrend_missing.png}\label{fig:detrend_w_missing}} 
%     \vspace{-0.3cm}
%     \caption{Trend extractors in missing and outlier case.}
%     \vspace{-0.3cm}
%     \label{fig:detrend}
% \end{figure}

\begin{figure}[!t]
    \centering
    \includegraphics[width=0.45\textwidth]{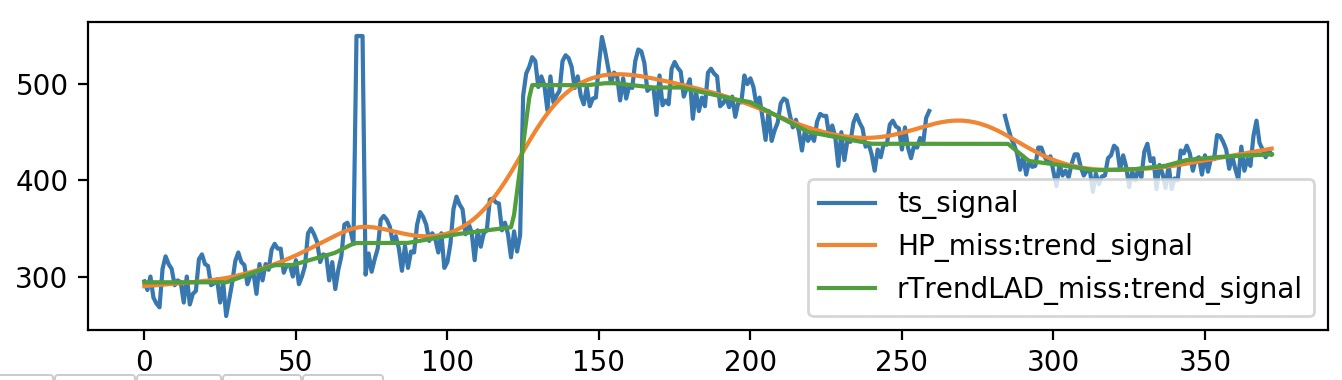}
    \vspace{-0.4cm}
    \caption{{Proposed robust trend filter without imputation vs. HP trend filter with linear interpolation under missing data, trend changes, and outliers (better in colorful version).}}
    \vspace{-2mm}
    \label{fig:detrend}
\end{figure}

Firstly, we carry out experiments to demonstrate the advantages of the designed robust trend filter under missing data. 
Fig.~\ref{fig:detrend} depicts the proposed robust trend filter without imputation versus common Hodrick–Prescott (HP) trend filter~\cite{hodrick1997postwar} with linear interpolation for time series under missing data, trend changes, and outliers. 
As HP filter is unable to deal with time series with missing data,
we inpaint the time series using linear interpolation and input it
into HP filter. After imputation, the HP filter is still biased by the outlier and abrupt changes of trend. In contrast, our method (denoted as "rTrendLAD\_miss") can deal with missing data without imputation, and meanwhile is robust to outliers and trend changes.

\begin{figure}[!t]
    \centering
    % \subfigure[Original time series ]{\includegraphics[width=0.3\textwidth]{plots/ori_p}\label{fig:ori_season}}\\
    \subfigure[\footnotesize{Linear interpolated time series}]{\includegraphics[width=0.35\textwidth]{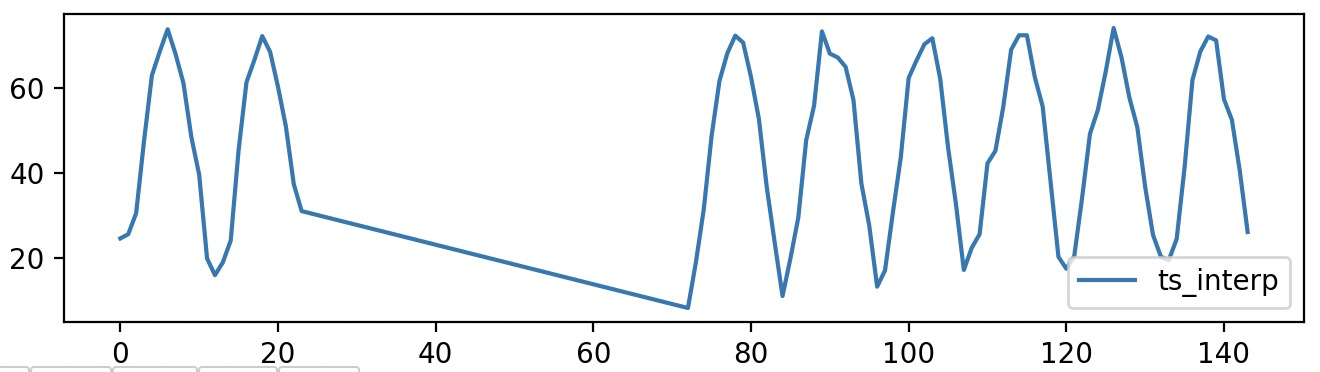}\label{fig:season_interp}}\vspace{-0.2cm}\\
    \subfigure[\footnotesize{Proposed robust ACF under missing data vs. ACF by linear interpolation (better in colorful version).}]{\includegraphics[width=0.35\textwidth]{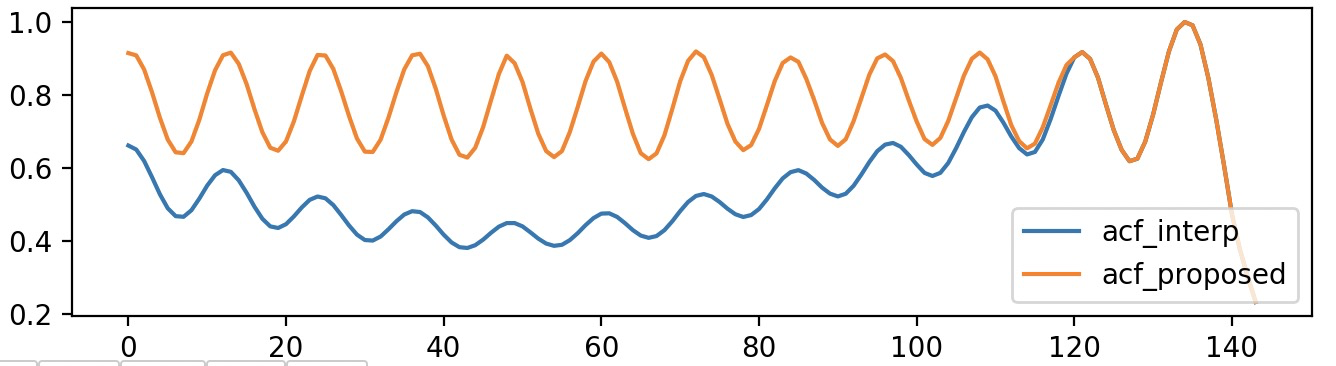}\label{fig:ACF_compare}}
    \vspace{-4mm}
    \caption{{Proposed robust ACF without imputation vs. ACF with linear interpolation under missing data.}}
    \vspace{-0.4cm}
    \label{fig:ACF}
\end{figure}

Next, we demonstrate the superiority of the
proposed robust ACF estimator under missing data. 
We conduct experiments on a time
series of length 144 with period 12.
We set the 24th to 71th points as missing, so the missing ratio is $1/3$.
Fig.~\ref{fig:season_interp} shows the incomplete time series with linear interpolation.
Fig.~\ref{fig:ACF_compare} shows the results of the proposed ACF under incomplete time series and the standard ACF under linear interpolation.
% We complete the time series using linear interpolation
% and compute its ACF as baseline. 
% We estimate the ACF 
% using the proposed method with incomplete time series.
% The ACF estimated using the proposed method with
% the incomplete series, and the normalized ACF of the linear interpolated time
% series, are illustrated in Fig.~\ref{fig:ACF_compare}.
We can see that
the ACF from linear interpolated time
series is severely affected, which makes it hard to find a threshold
to locate the peaks of the ACF. In contrast, our proposed robust ACF estimator can
provide a promising ACF estimation which brings accurate periodicity detection.

%% file: 5_conclusion.tex
\vspace{-2mm}
\section{Conclusion and Future Work}
\vspace{-3mm}

In this paper we propose a novel periodicity detection algorithm to detect the dominant periodicity. Our algorithm can robustly deal with trend changes and outliers for time series under missing data scenarios. In the future we plan to extend our framework to handle time series with multiple periodicities, as well as apply it in more real-world applications.

% By formulating the estimation of periodogram as a regression problem, we propose a robust M-periodogram based on Huber loss and then apply Fisher's test to determine if there is a dominant periodicity. Compared with other periodicity detection algorithms, our algorithm can handle missing values properly. 

% We rigorous prove that our algorithm can still work successfully when the length of the missing block is less than $1/3$ of the period length. 

%% file: 0_main.bbl
\begin{thebibliography}{10}

\bibitem{kim2020periodicity}
Heonho Kim, Unil Yun, Bay Vo, Jerry Chun-Wei Lin, and Witold Pedrycz,
\newblock ``Periodicity-oriented data analytics on time-series data for
  intelligence system,''
\newblock {\em IEEE Systems Journal}, vol. 15, no. 4, pp. 4958--4969, 2020.

\bibitem{wen2022kddtimeseries}
Qingsong Wen, Linxiao Yang, Tian Zhou, and Liang Sun,
\newblock ``Robust time series analysis and applications: An industrial
  perspective,''
\newblock in {\em KDD 2022}, 2022, pp. 4836--4837.

\bibitem{zhang2020anomaly}
Shuo Zhang, XiaoFei Chen, JiaYuan Chen, Qiao Jiang, and Hejiao Huang,
\newblock ``Anomaly detection of periodic multivariate time series under high
  acquisition frequency scene in {IoT},''
\newblock in {\em Int. Conf. on Data Mining Workshops}, 2020, pp. 543--552.

\bibitem{jingkun20_TAD}
Jingkun Gao, Xiaomin Song, Qingsong Wen, Pichao Wang, Liang Sun, and Huan Xu,
\newblock ``{RobustTAD}: Robust time series anomaly detection via decomposition
  and convolutional neural networks,''
\newblock {\em KDD Workshop MileTS}, 2020.

\bibitem{tolas2021periodicity}
Ramona Tolas, Raluca Portase, Andrei Iosif, and Rodica Potolea,
\newblock ``Periodicity detection algorithm and applications on {IoT} data,''
\newblock in {\em 20th International Symposium on Parallel and Distributed
  Computing}, 2021, pp. 81--88.

\bibitem{zhang2022tfad}
Chaoli Zhang, Tian Zhou, Qingsong Wen, and Liang Sun,
\newblock ``{TFAD}: A decomposition time series anomaly detection architecture
  with time-frequency analysis,''
\newblock in {\em CIKM}, 2022.

\bibitem{WenRobustPeriod20}
Qingsong Wen, Kai He, Liang Sun, Yingying Zhang, Min Ke, and Huan Xu,
\newblock ``{RobustPeriod}: Time-frequency mining for robust multiple
  periodicity detection,''
\newblock in {\em ACM Int. Conf. on Management of Data (SIGMOD)}, 2021, pp.
  2328--2337.

\bibitem{xu2021two}
Qingyang Xu, Qingsong Wen, and Liang Sun,
\newblock ``Two-stage framework for seasonal time series forecasting,''
\newblock in {\em ICASSP}, 2021, pp. 3530--3534.

\bibitem{zhou2022fedformer}
Tian Zhou, Ziqing Ma, Qingsong Wen, Xue Wang, Liang Sun, and Rong Jin,
\newblock ``Fedformer: Frequency enhanced decomposed transformer for long-term
  series forecasting,''
\newblock in {\em International Conference on Machine Learning}, 2022, pp.
  27268--27286.

\bibitem{hyndman2006:characterizeTS}
X.~Wang, K.~Smith-Miles, and R.~Hyndman,
\newblock ``Characteristic-based clustering for time series data,''
\newblock {\em Data Mining and Knowledge Discovery}, vol. 13, pp. 335--364, 09
  2006.

\bibitem{vlachos2004identifying}
Michail Vlachos, Christopher Meek, Zografoula Vagena, and Dimitrios Gunopulos,
\newblock ``Identifying similarities, periodicities and bursts for online
  search queries,''
\newblock in {\em ACM SIGMOD Int. Conf. on Management of data}, 2004, pp.
  131--142.

\bibitem{STL_cleveland1990stl}
Robert~B Cleveland, William~S Cleveland, Jean~E McRae, and Irma Terpenning,
\newblock ``{STL}: A seasonal-trend decomposition procedure based on loess,''
\newblock {\em Journal of Official Statistics}, vol. 6, no. 1, pp. 3--73, 1990.

\bibitem{wen2019robuststl}
Qingsong Wen, Jingkun Gao, Xiaomin Song, Liang Sun, Huan Xu, and Shenghuo Zhu,
\newblock ``{RobustSTL}: A robust seasonal-trend decomposition algorithm for
  long time series,''
\newblock in {\em AAAI Conference on Artificial Intelligence}, 2019, pp.
  5409--5416.

\bibitem{yang2021robuststl}
Linxiao Yang, Qingsong Wen, Bo~Yang, and Liang Sun,
\newblock ``A robust and efficient multi-scale seasonal-trend decomposition,''
\newblock in {\em ICASSP}, 2021.

\bibitem{yi2016st}
Xiuwen Yi, Yu~Zheng, Junbo Zhang, and Tianrui Li,
\newblock ``{ST-MVL}: filling missing values in geo-sensory time series data,''
\newblock in {\em Int. Joint Conf. on Artificial Intelligence (IJCAI)}, 2016.

\bibitem{fisher1929tests}
Ronald~Aylmer Fisher,
\newblock ``Tests of significance in harmonic analysis,''
\newblock {\em Proceedings of the Royal Society of London. Series A, Containing
  Papers of a Mathematical and Physical Character}, vol. 125, no. 796, pp.
  54--59, 1929.

\bibitem{wichert2004identifying}
S.~Wichert, K.~Fokianos, and K.~Strimmer,
\newblock ``Identifying periodically expressed transcripts in microarray time
  series data,''
\newblock {\em Bioinformatics}, vol. 20, no. 1, pp. 5--20, 2004.

\bibitem{Wang2006}
{Wang, J.}, {Chen, T.}, and {Huang, B.},
\newblock ``Cyclo-period estimation for discrete-time cyclo-stationary
  signals,''
\newblock {\em IEEE Transactions on Signal Processing}, vol. 54, no. 1, pp.
  83--94, 2006.

\bibitem{Toller2019}
Maximilian Toller, Tiago Santos, and Roman Kern,
\newblock ``{SAZED: parameter-free domain-agnostic season length estimation in
  time series data},''
\newblock {\em Data Mining and Knowledge Discovery}, 2019.

\bibitem{cPD_vlachos2005Autoperiod}
Michail Vlachos, Philip Yu, and Vittorio Castelli,
\newblock ``On periodicity detection and structural periodic similarity,''
\newblock in {\em SIAM Int. Conf. on Data Mining}, 2005, pp. 449--460.

\bibitem{almasri2011new}
Abdullah Almasri,
\newblock ``A new approach for testing periodicity,''
\newblock {\em Communications in Statistics—Theory and Methods}, vol. 40, no.
  7, pp. 1196--1217, 2011.

\bibitem{hu2014periodicity}
Feiyan Hu, Alan~F Smeaton, and Eamonn Newman,
\newblock ``Periodicity detection in lifelog data with missing and irregularly
  sampled data,''
\newblock in {\em IEEE BIBM}, 2014, pp. 16--23.

\bibitem{glynn2006detecting}
Earl~F Glynn, Jie Chen, and Arcady~R Mushegian,
\newblock ``{Detecting periodic patterns in unevenly spaced gene expression
  time series using Lomb--Scargle periodograms},''
\newblock {\em Bioinformatics}, vol. 22, no. 3, pp. 310--316, 2006.

\bibitem{lomb1976least}
Nicholas~R Lomb,
\newblock ``Least-squares frequency analysis of unequally spaced data,''
\newblock {\em Astrophysics and space science}, vol. 39, no. 2, pp. 447--462,
  1976.

\bibitem{wen2019robusttrend}
Qingsong Wen, Jingkun Gao, Xiaomin Song, Liang Sun, and Jian Tan,
\newblock ``{RobustTrend}: a {Huber} loss with a combined first and second
  order difference regularization for time series trend filtering,''
\newblock in {\em IJCAI}, 2019, pp. 3856--3862.

\bibitem{boyd}
Stephen Boyd, Neal Parikh, Eric Chu, Borja Peleato, Jonathan Eckstein, et~al.,
\newblock ``Distributed optimization and statistical learning via the
  alternating direction method of multipliers,''
\newblock {\em Foundations and Trends in Machine learning}, vol. 3, no. 1, pp.
  1--122, 2011.

\bibitem{Wiener1930}
Norbert Wiener,
\newblock ``{Generalized harmonic analysis},''
\newblock {\em Acta Math.}, vol. 55, pp. 117--258, 1930.

\bibitem{Scholkmann2012}
Felix Scholkmann, Jens Boss, and Martin Wolf,
\newblock ``An efficient algorithm for automatic peak detection in noisy
  periodic and quasi-periodic signals,''
\newblock {\em Algorithms}, vol. 5, no. 4, pp. 588--603, 2012.

\bibitem{cran_hornik2012comprehensive}
Kurt Hornik,
\newblock ``The comprehensive {R} archive network,''
\newblock {\em Wiley interdisciplinary reviews: Computational statistics}, vol.
  4, no. 4, pp. 394--398, 2012.

\bibitem{hodrick1997postwar}
Robert~J Hodrick and Edward~C Prescott,
\newblock ``Postwar {US} business cycles: an empirical investigation,''
\newblock {\em Journal of Money, Credit, and Banking}, pp. 1--16, 1997.

\end{thebibliography}
